\documentclass{article}

\PassOptionsToPackage{numbers, compress}{natbib}

\usepackage[preprint]{neurips_2020}

\usepackage[utf8]{inputenc} 
\usepackage[T1]{fontenc}    
\usepackage{hyperref}       
\usepackage{url}            
\usepackage{booktabs}       
\usepackage{amsfonts}       
\usepackage{nicefrac}       
\usepackage{microtype}      
\usepackage{multirow}       
\usepackage{makecell}       
\usepackage{graphicx}       
\usepackage{amsbsy}         

\title{Identifying and Characterising Response in Clinical Trials: Development and Validation of a Machine Learning Approach in Colorectal Cancer}

\author{%
  Adam P. Marcus \\
  Department of Computing \\
  Imperial College London \\
  London, SW7 2AZ, UK \\
  \texttt{adam.marcus11@imperial.ac.uk} \\
  \And
  Paul Agapow \\
  Oncology R\&D ML\&AI \\
  AstraZeneca \\
  Cambridge, CB2 1PQ, UK \\
  \texttt{paul.agapow@astrazeneca.com} \\
}

\begin{document}

\maketitle

\begin{abstract}
  Precision medicine promises to transform health care by offering individualised treatments that dramatically improve clinical outcomes. A necessary prerequisite is to identify subgroups of patients who respond differently to different therapies. Current approaches are limited to static measures of treatment success, neglecting the repeated measures found in most clinical trials. Our approach combines the concept of partly conditional modelling with treatment effect estimation based on the Virtual Twins method. The resulting time-specific responses to treatment are characterised using survLIME, an extension of Local Interpretable Model-agnostic Explanations (LIME) to survival data. Performance was evaluated using synthetic data and applied to clinical trials examining the effectiveness of panitumumab to treat metastatic colorectal cancer. An area under the receiver operating characteristic curve (AUC) of 0.77 for identifying fixed responders was achieved in a 1000 patient simulation. When considering dynamic responders, partly conditional modelling increased the AUC from 0.597 to 0.685. Applying the approach to colorectal cancer trials found genetic mutations, sites of metastasis, and ethnicity as important factors for response to treatment. Our approach can accommodate a dynamic response to treatment while potentially providing better performance than existing methods in instances of a fixed response to treatment. When applied to clinical data we attain results consistent with the literature.
\end{abstract}

\section{Introduction}
Precision medicine is set to become the new paradigm of health care and revolutionise drug development\citep{RN132, RN133}. The promise is that diagnostic tools may integrate individual patient characteristics with knowledge of underlying disease pathophysiology to allow matching to safe and effective drugs. Creation of such tools is becoming increasingly feasible owing to the expanding volume of health data combined with recent innovations in artificial intelligence. In particular, the management of cancer is expected to benefit greatly from this progress and in recent years has already seen a number of tailored therapies become essential treatments\citep{RN35, RN36}. At the same time, it has become clear that patients may display a widely varying response to the same therapies. As a consequence, modern drug development must no longer consider the population as a homogeneous whole but looks to find patient subgroups, defined by individual characteristics, that show the highest positive and negative treatment effects\citep{RN34}. The problem of determining these subgroups with differential treatment effects is sometimes referred to as \textit{responder identification} but is more widely known as \textit{subgroup identification}\citep{RN21}.

Many different subgroup identification methods have been developed. The classical approach is well described by Kehl and Ulm\citep{RN1} and includes fitting a model containing treatment-covariate interactions. To address the shortcomings of this approach several techniques have been developed around the classification and regression tree methodology\citep{RN41}. These include techniques that directly explore the relationship between treatment effects and covariates such as Interaction Trees\citep{RN42}, Virtual Twins \citep{RN21}, and Subgroup Identification Based on Differential Effect Search (SIDES) \citep{RN22}. Also, techniques that indirectly explore this relationship such as 
Qualitative Interaction Trees \citep{RN47} and Generalised, Unbiased, Interaction Detection and Estimation (GUIDE) \citep{RN49}. Non-tree methods, while less popular, have also been considered\citep{RN141}. The most notable of these techniques are penalised regression models although Bayesian approaches have also been proposed\citep{RN142, RN143}. One area frequently encountered in real world data but seldom addressed by these methods is the handling of repeated time dependent observations. In this study we aim to: (i) develop an approach using repeated measures to identify and characterise responders in clinical trial data; (ii) validate the approach using simulated data; and (iii) apply the approach to colorectal cancer chemotherapy trial data.

\section{Methods}

\subsection{Responder identification and characterisation}
\begin{figure}
  \centering
  \includegraphics{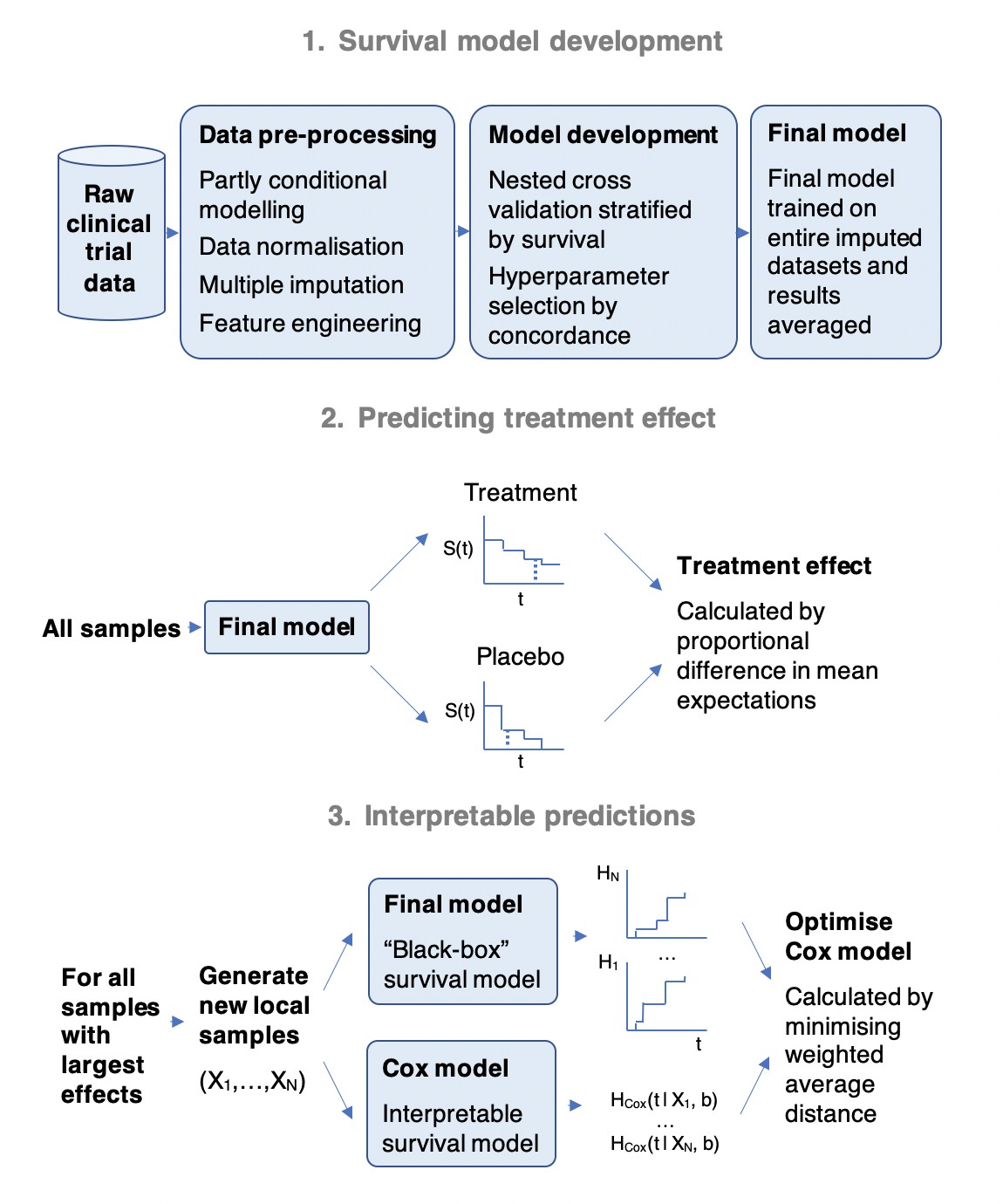}
  \caption{A schematic illustration of the main stages of the proposed approach.}
  \label{fig:approach}
\end{figure}

The proposed approach is illustrated in Figure \ref{fig:approach}. The concept is based upon the Virtual Twins method where responders are identified by their individual treatment effect difference. For each patient, only the outcome to one of the two possible treatment assignments is observed, however by predicting the outcome under the other assignment, it is possible to calculate the treatment effect difference. The challenge of identifying responders then becomes one of building the best predictive model. Therefore, in the first stage, the proposed approach borrows ideas from data science to train a high performing predictive model. In the second stage, the individual treatment effects are estimated. In the final stage, responders are characterised by interpreting the individual model predictions.

\subsubsection{Survival model development}
\paragraph{Data pre-processing}
To account for time-varying covariates, a new dataset is generated using the principles of partly conditional models\citep{RN12}. This involves treating each set of covariates at a given time as a separate individual, setting the event times to the residual time-to-event together with adjusting the censoring indicator. Covariates are encoded into numerical representations and normalised to facilitate model fitting. Missing values are imputed with the same method as van Houwelingen and Putter\citep{RN77} by carrying forward the last observed value for each patient’s covariates. Multiple imputation\citep{RN136} is used to accommodate missing values in baseline measurements. All variables with a constant variance are removed and a stepwise selection procedure is used to remove covariates with a variance inflation factor above 5 using the method described by Craney and Surles\citep{RN86}. All patients with an event too soon after recruitment are removed. In the clinical study a cut-off of 14 days after initiation of treatment was used to match the length of a single cycle of chemotherapy. Similarly, all sets of covariates measured too close to a subsequent event are removed to prevent possible leakage of information about the event.

\paragraph{Model development}
The pre-processed data is used to train and evaluate different candidate algorithms. A penalised Cox model serving as a baseline incorporating only linear and time-invariant effects of covariates. Random survival forests and DeepSurv\citep{RN13} selected as recently developed methods to accommodate non-linear and time-dependent effects of covariates\citep{RN14, RN13}. Lastly, WTTE-RNN\citep{RN17} chose due to its ability to handle non-linear time-dependent effects and utilise previous historical observations when making its predictions. To minimise potential bias introduced by using the same data to develop and validate a model, a nested cross-validation scheme\citep{RN80} is employed with 10-fold inner loops and 10-fold outer loops. Cross-validation is stratified by both time-to-event and censoring status, and is grouped by patient to ensure that observations at different times do not appear in both the training and testing sets. Evaluation of model performance is achieved using the averaged time-dependent concordance index\citep{RN9}. To account for partly conditional modelling (PCM), a combined survival function is produced for each patient consisting of the predictions at each different time provided. Time-dependent concordance index was chosen as it is an extension of Harrell’s c-index to use the whole predicted survival function\citep{RN9}.

\paragraph{Final model creation}
To create the final model, the cross-validation inner loops procedure, including hyperparameter estimation, is repeated over the whole dataset using the best selected algorithm. 

\subsubsection{Predicting treatment effects}
The final model is used to make predictions for the unobserved treatment assignment of every patient at every given time in the dataset. The proportional change between the estimated prediction times under one treatment assignment and the actually observed times under the other treatment assignment is calculated. The treatment effect is then defined as the log of the proportional change for each patient at a given time. On this scale, a value greater than zero would suggest being a responder, a value equal to zero a non-responder, and a value less than zero an anti-responder.

\subsubsection{Interpretable predictions}
A representative sample of responders, non-responders, and anti-responders must be obtained. Assuming that treatment effects are sorted in ascending order, the first sampling region is the lower quartile of all anti-responders; the second sampling region encompasses the upper quartile of all anti-responders, all non-responders, and the lower quartile of all responders; and the last sampling region is the upper quartile of all responders. For every patient with specific times in the responder region, the survLIME-Inf technique is used. Rather than in the original algorithm sampling all covariates uniformly around the point of interest, only certain variables are sampled. This is necessary to provide a means to restrict the region of covariate space approximated by the surrogate model allowing for a better approximation. Consequently, the variables representing treatment and time since recruitment are kept constant as only factors relating to the response to this treatment at a specific time is of interest. This produces a list of log hazard ratios (HR) for every time the technique is applied that can then be averaged to give the important factors for response to treatment. 

\subsection{Validation using simulated data}
A simulation study was performed using an experimental setup similar to that used by Foster et al. when developing the Virtual Twins method. Notably, the simulation was extended to support time-varying covariates. The case of a 2-arm randomised control trial was considered. The data consisted of $(Y_{i,j},D_{i,j},X_{1,i,j},...,X_{p,i,j})$, $i=t_1,...,t_n$ and $j=1,...,m$ where $i$ indicates the observation time, $j$ the patient identifier, $p$ the number of covariates, $Y$ the time-to-event outcome, and $D$ the treatment indicator. Let $m_j=$ number of observations with $D_i=j$ when $j=0,1$. It is expected that $m_0\approx m_1$ in a 1:1 clinical trial. The goal was to identify a region representing responders to treatment (denoted $A$) of the covariate space of $X$. Let $\hat{A}$ be the estimated region determined by the treatment effect described in the proposed approach above. A base-case simulation was defined followed by a number of modifications.

\subsubsection{Fixed responder base case}
In the base case, 1000 patients were simulated. The $X$’s were all generated as independent. Variables $X_1$ to $X_6$ were time-invariant and followed a normal distribution $X_{j,t_1}... X_{j, t_n} \sim \mathcal{N}\left(0,1\right), j=1, ...,6$. Variables $X_7$ to $X_{15}$ were time-varying and were produced by an autoregressive-moving-average (ARMA) process\citep{RN89} with parameters $\varphi_1=1$, $\theta_1=1$, $\theta_2=1$, and a normally distributed constant term $X_{j,t}=\mathcal{N}\left(0,1\right)+\varepsilon_t+\sum_{i=1}^{p}\varphi_iX_{j,t-1}+\sum_{i=1}^{q}\theta_i\varepsilon_{t-1}$, $j=7,...,15$. An ARMA process was chosen due to its simplicity and as it has previously been shown to be successful at modelling biomedical time series\citep{RN90}. The survival times were generated following a Cox proportional hazards model using the method described by Hendry\citep{RN4} with a baseline cumulative hazard function $=t^2$. The hazard function was set to $1 - 0.5X_1 - 0.5X_2 + 0.5X_7 - 0.1D - 0.5X_2X_7 - \theta DI (\boldsymbol{X}\in A)$, where $\theta$ represented the degree to which patients in region $A$ have an enhanced treatment effect. The region $A$ was defined as $X_1>0 \cap X_2<0$, which implied that approximately a quarter of the patients were responders to treatment. The observation times considered were $t_1, ..., t_n=\left\{0, 0.1, 0.2, ... 10\right\}$. Performance was evaluated on 100 repeats of the simulation. In the base case $\theta=0.9$. A null fixed responder case was also considered where there is no enhanced treatment effect, $\theta=0$. For both the base and null case, the proposed approach was compared with and without PCM to account for time-varying covariates.

\subsubsection{Dynamic responder case and extensions}
The base case was extended to assume that variables in the definition of region $A$ were no longer fixed but allowed to vary in time. A clinical justification for such a phenomenon is temporal intratumour heterogeneity, that describes tumour diversity arising from genetic changes over time. In the simulation, the values of $X_1$ and $X_2$ were changed once, at an uninformative, independently chosen time $m=\mathcal{U}(0,1)$, and were assigned new values from the normal distribution therefore $X_{j,t_1}... X_{j, t_m} \sim \mathcal{N}\left(0,1\right)$ and $X_{j,t_m}... X_{j, t_n} \sim \mathcal{N}\left(0,1\right)$ where $j=1,2$. This resulted in the values of $X_1$ and $X_2$ changing for approximately half of patients. The dynamic responder case was further extended by considering when: (i) the number of covariates was doubled from 15 to 30 to better represent the amount of data found in real clinical trials; and (ii) the number of simulated patients was set to either 300 or 2000 to mirror the number of participants found in phase 2 and 3 drug trials respectively. 

\subsubsection{Evaluating performance}
A number of different criteria were used to evaluate the performance of the proposed approach. Firstly, to assess the closeness of the estimated region $\hat{A}$ with the true region $A$, the following metrics were used: sensitivity, specificity, positive predictive value (PPV), negative predictive value (NPV), and area under the receiver operating characteristic curve (AUC). Sensitivity, specificity, PPV, and NPV were calculated by applying a threshold $c$ to the treatment effect scores produced by the proposed approach. In this study, a value of $c=0.15$ was used for illustrative purposes. By varying $c$, receiver operating characteristic (ROC) curves and AUC scores were calculated using knowledge of the true $A$ membership. Confidence bands for the ROC curves in this study were determined using the fixed-width bands technique\citep{RN130}. In the null case, when $\theta=0$, region $A$ does not exist therefore only specificity carried meaning and for a perfect prediction would equal 1. Secondly, to assess whether the correct covariates were identified that describe region $A$, the following metrics were constructed. The HR of the important factors produced by the proposed approach were sorted by largest to smallest effect size. The number of times the factors defining region $A$, $X_1$ and $X_2$, were included in the top two positions of the list were counted. Additionally, the mean of the absolute log HR of these top two positions were calculated. This metric represented the average magnitude of the effect and was useful at evaluating the null case where an ideal value would be 0.

\subsection{Application to clinical trial data}
The raw data were obtained from Project Data Sphere\citep{RN65} for four colorectal cancer trials\citep{RN29, RN30, RN10, RN32}. The primary data for these trials was collected to test the effectiveness of panitumumab combined with varying chemotherapy agents to treat metastatic colorectal cancer. The proposed approach was then applied to each of the studies independently resulting in HR of disease progression for each variable that are important to responders to treatment. The findings were then assessed for plausibility and compared with the literature.

\section{Results}
\subsection{Validation using simulated data}

\begin{figure}
  \centering
  \includegraphics{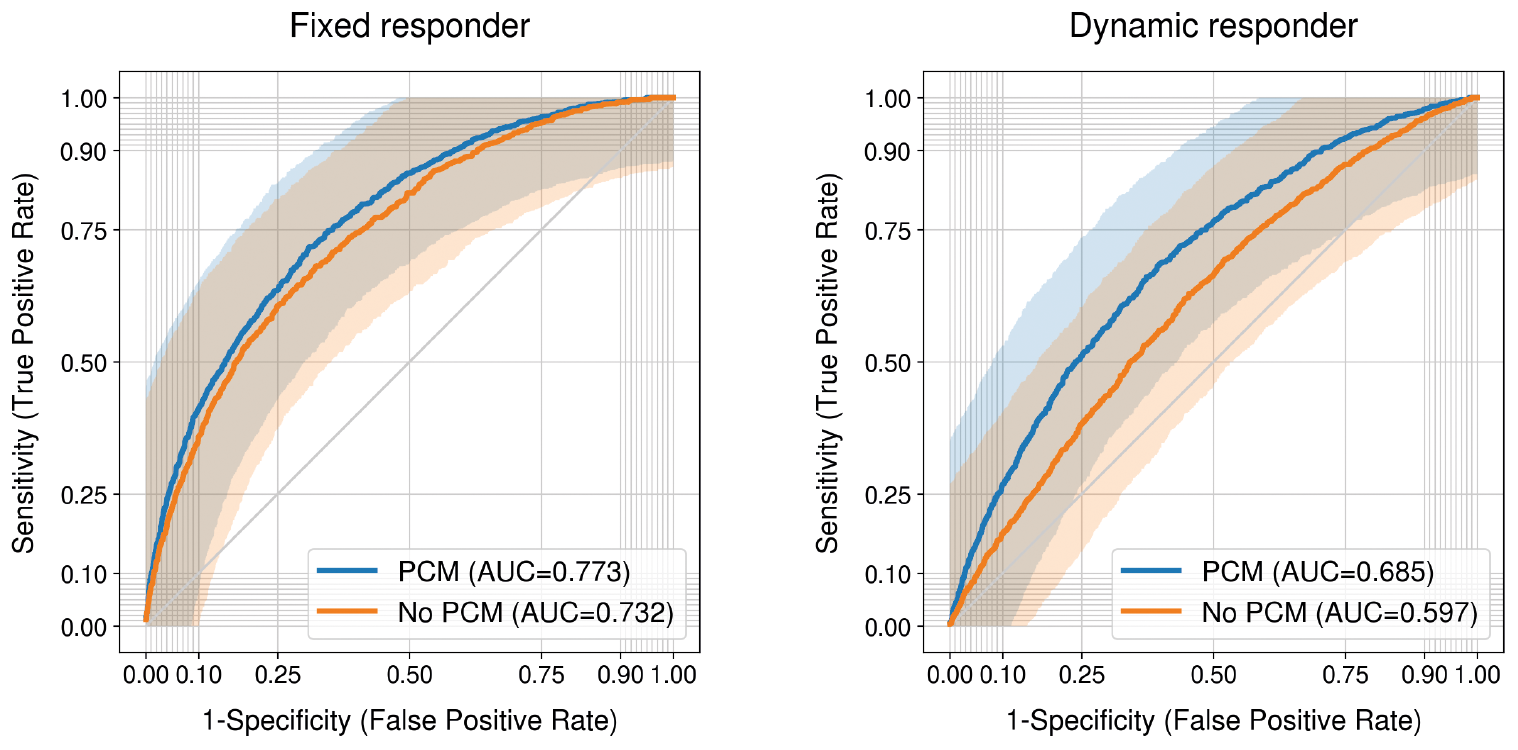}
  \caption{Receiver operating characteristic curves for identification performance in both the fixed and dynamic responder simulation. Shaded areas represent the 95\% confidence bands.}
  \label{fig:roc-curve}
\end{figure}

The hyperparameter search found the DeepSurv algorithm with two layers achieved the highest concordance index. As shown in Tables \ref{table:fixed-case} and \ref{table:dynamic-case}, the performance of the proposed approach improved when applying PCM. Evaluating identification performance, AUC increased from 0.732 to 0.773 in the fixed case and from 0.597 to 0.685 in the dynamic case. Characterisation performance also improved with the correct factors being found more often (top two HR count 1.117 vs 0.979 and 1.032 vs 0.830). However, when comparing the null cases, the approach with PCM has lower specificity (0.874 vs 0.991 and 0.907 vs 0.997) and higher measured effect size (mean absolute log HR of 0.258 vs 0.251 and 0.242 vs 0.234). Identification performance of the approach was worse in the dynamic responder case than the fixed responder base case as shown in Figure \ref{fig:roc-curve}. Larger sample sizes benefit the proposed approach as illustrated in Table \ref{table:extended-cases}. Doubling the sample size increased the AUC from 0.685 to 0.742 with PCM and from 0.597 to 0.652 without. Doubling the number of covariates led to a slight reduction in AUC from 0.685 to 0.681 with PCM and from 0.597 to 0.578 without. These results suggest that, given sufficient sample size, the proposed approach can provide satisfactory performance at responder identification and characterisation. Furthermore, that the use of PCM increases performance with the possible trade-off of a small increase in the chance of type I errors.

\begin{table}
  \caption{Evaluating performance in the fixed responder base case simulation. }
  \label{table:fixed-case}
  \centering
  \setlength{\tabcolsep}{5pt} 
  \begin{tabular}{rrccccccc}
    \toprule 
      & & \multicolumn{5}{c}{Identification} & \multicolumn{2}{c}{Characterisation} \\
      \cmidrule(r){3-9}
      & & AUC & Sensitivity	& Specificity &	PPV & NPV & \makecell{Mean\\|Log(HR)|} & \makecell{Top 2 HR\\count} \\
      \midrule
      \multirow{2}{3.11em}{$\theta = 0.9$} & No PCM & 0.732 & 0.792 & 0.543 & 0.362 & 0.886 & 0.301 & 0.979 \\
      & PCM & 0.773 & 0.824 & 0.577 & 0.380 & 0.905 & 0.321 & 1.117 \\
      \multirow{2}{3.11em}{$\theta = 0$} & No PCM & - & - & 0.991 & - & - & 0.251 & 0.718 \\
      & PCM & - & - & 0.874 & - & - & 0.258 & 0.628 \\
    \bottomrule
  \end{tabular}
\end{table}

\begin{table}
  \caption{Evaluating performance in the dynamic responder case simulation. }
  \label{table:dynamic-case}
  \centering
  \setlength{\tabcolsep}{5pt} 
  \begin{tabular}{rrccccccc}
    \toprule 
      & & \multicolumn{5}{c}{Identification} & \multicolumn{2}{c}{Characterisation} \\
      \cmidrule(r){3-9}
      & & AUC & Sensitivity	& Specificity &	PPV & NPV & \makecell{Mean\\|Log(HR)|} & \makecell{Top 2 HR\\count} \\
      \midrule
      \multirow{2}{3.11em}{$\theta = 0.9$} & No PCM & 0.597 & 0.478 & 0.660 & 0.310 & 0.787 & 0.256 & 0.830 \\
      & PCM & 0.685 & 0.680 & 0.591 & 0.360 & 0.849 & 0.297 & 1.032 \\
      \multirow{2}{3.11em}{$\theta = 0$} & No PCM & - & - & 0.997 & - & - & 0.234 & 0.381 \\
      & PCM & - & - & 0.907 & - & - & 0.242 & 0.371 \\
    \bottomrule
  \end{tabular}
\end{table}

\begin{table}
  \caption{Evaluating performance in modifications to the dynamic responder case simulation. The number of patients ($N$) is varied as is the number of covariates ($P$).}
  \label{table:extended-cases}
  \centering
  \setlength{\tabcolsep}{4.8pt} 
  \begin{tabular}{rrccccccc}
    \toprule 
      & & \multicolumn{5}{c}{Identification} & \multicolumn{2}{c}{Characterisation} \\
      \cmidrule(r){3-9}
      & & AUC & Sensitivity	& Specificity &	PPV & NPV & \makecell{Mean\\|Log(HR)|} & \makecell{Top 2 HR\\count} \\
      \midrule
      \multirow{2}{4.3em}{$N = 300$} & No PCM & 0.547 & 0.420 & 0.617 & 0.272 & 0.762 & 0.248 & 0.500 \\
      & PCM & 0.605 & 0.637 & 0.536 & 0.318 & 0.810 & 0.259 & 1.872 \\
      \multirow{2}{4.3em}{$N = 1000$} & No PCM & - & - & 0.997 & - & - & 0.234 & 0.381 \\
      & PCM & - & - & 0.907 & - & - & 0.242 & 0.371 \\
      \multirow{2}{4.3em}{$N = 2000$} & No PCM & 0.597 & 0.478 & 0.660 & 0.310 & 0.787 & 0.256 & 0.830 \\
      & PCM & 0.685 & 0.680 & 0.591 & 0.360 & 0.849 & 0.297 & 1.032 \\
      \multirow{2}{4.3em}{$P = 30$} & No PCM & - & - & 0.997 & - & - & 0.234 & 0.381 \\
      & PCM & - & - & 0.907 & - & - & 0.242 & 0.371 \\
    \bottomrule
  \end{tabular}
\end{table}

\subsection{Application to clinical trial data}

\begin{figure}
  \centering
  \includegraphics{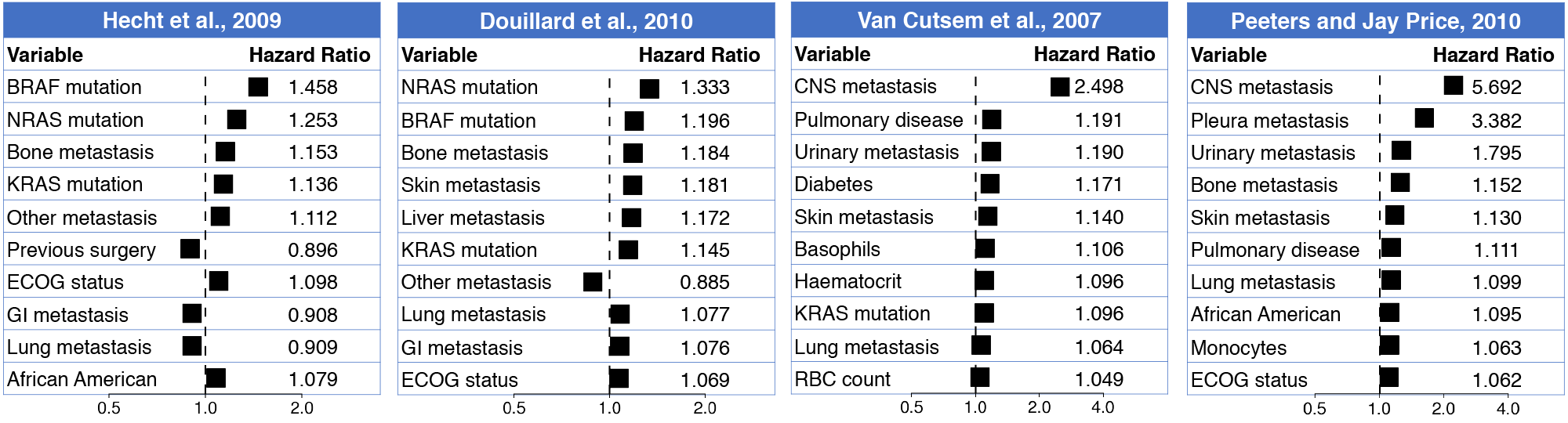}
  \caption{Results of the approach applied to the four clinical studies.}
  \label{fig:clincal-results}
\end{figure}

For each clinical trial, the HR's of disease progression for the most important response to treatment factors are shown in Figure \ref{fig:clincal-results}. Mutations in the KRAS (Hecht et al., 2009, HR 1.136; Douillard et al., 2010, HR 1.145; Van Cutsem et al., 2007, HR 1.096), BRAF (Hecht et al., 2009, HR 1.458; Douillard et al., 2010, HR 1.196), and NRAS (Hecht et al., 2009, HR 1.253; Douillard et al., 2010, HR 1.333) genes were among the factors with the biggest effects across all trials. The distribution of metastatic spread and particularly spread to the central nervous system (CNS) (Van Cutsem et al., 2007, HR 2.498 and Peeters and Jay Price, 2010, HR 5.692), bone (Hecht et al., 2009, HR 1.153; Douillard et al., 2010, HR 1.184; and Peeters and Jay Price, 2010, HR 1.152), and skin (Douillard et al., 2010, HR 1.181; Van Cutsem et al., 2007, HR 1.140; and Peeters and Jay Price, 2010, HR 1.130) were found to be important. Lastly, Black or African American race was identified as key factors (Hecht et al., 2009, HR 1.079 and Peeters and Jay Price, 2010, HR 1.095).

\section{Discussion}
\subsection{Principal findings and interpretation}
Precision medicine promises to transform health care by utilising the increasing amounts of health data to offer individualised treatments that will dramatically improve clinical outcomes. Integral to realising this promise is the ability to accurately identify subgroups of patients who will benefit the most from a given treatment. To date, a number of different approaches have been developed that go beyond conventional pre-defined subgroups and instead mine clinical trial data retrospectively\citep{RN21, RN48, RN34, RN22, RN47}. Much of the existing literature has exclusively focused on baseline measurements taken at the beginning of the study. This overlooks repeated measurements that now routinely accompany the multi-year clinical trials required in drug development. By ignoring these measures, a potential opportunity is lost to find better subgroups provided by the additional data. Furthermore, an implicit assumption is being made about the nature of the response to treatment, that response status is fixed and not allowed to vary with time. The validity of this assumption is debatable, particularly in the context of cancer trials where the underlying disease is dynamic and can acquire traits over time\citep{RN94}. In the present study, an approach is proposed that relaxes this assumption by combining the concept of partly conditional modelling\citep{RN12} with treatment effect estimation based on the Virtual Twins method\citep{RN21}. This results in time-specific responses to treatment that are then characterised using the recently developed extension of the LIME technique to survival data, survLIME\citep{RN3, RN2}.

\subsection{Validation using simulated data}
As suggested by the results of the simulation study, whether considering a fixed or variable response status, the proposed approach appears to benefit from the application of PCM. While not directly comparable, the approach outlined in this study achieves similar identification performance, with a fixed responder base case AUC of 0.773 compared to 0.77, to that described by Foster et. al\citep{RN21} for an inherently harder task – as half the covariates including one involved in the hazard function are allowed to vary in time. It is reasonable to expect that in an identical fixed responder simulation the proposed approach could achieve superior performance. Objective comparison of characterisation performance with other methods is difficult given the different outcomes produced by the proposed approach and consequently the different metrics used in this study.

The simulation study also explored the effects of varying sample size with the proposed approach. In drug development, clinical trials normally proceed through several phases with an increasing number of participants tested over increased spans of time to reveal less common side effects\citep{RN107}. Results of the simulation with 300 patients, representative of a large phase 2 clinical trial, suggested that even if assuming a fixed response status, identifying the correct responders is unlikely. To achieve reasonable results, larger sample sizes in the region of greater than 1000 patients are required, particularly if responder status is dynamic. Typically, such sample sizes are exclusive to event-driven phase 3 clinical drug trials potentially limiting the effectiveness of this technique. Although the relation between performance and sample size is not unique to the proposed approach and has been observed in other responder identification techniques\citep{RN21, RN108, RN109}.

Clinical trials often investigate a wide variety of potential biomarkers, considerably more than the 15 covariates used in the simulated base case. There is a clear advantage for a technique capable of accommodating these additional covariates in order to increase the chances of including the correct factors needed to accurately explain responders to treatment. As previously shown, doubling the number of covariates with the proposed approach led to only a small decrease in identification performance (0.685 to 0.681 AUC). This suggests that the approach may be viable with increased numbers of covariates although should perhaps be used with caution if applied to trials with hundreds or more covariates. In actual trials, it is often the case that certain factors may have an a priori biological rationale for inclusion or exclusion. Therefore, mitigation could be provided by manual selection of these factors.

\subsection{Application to clinical trial data}
Applying the proposed approach to colorectal cancer trials investigating the effectiveness of panitumumab found genetic mutations, sites of metastasis, and ethnicity as consistently important factors influencing response to treatment. Much research has focused on the significance of genetic mutations in the management of metastatic colorectal cancer\citep{RN110, RN111, RN112}. It is therefore reassuring that KRAS, NRAS, and BRAF mutations were identified as important as they have already been shown to be negative predictive biomarkers for panitumumab therapy\citep{RN114, RN116, RN117, RN118}. All three of the sites of spread identified by the proposed approach are known to carry a poor prognosis. The selection of these factors may arise from the fact that there was no histological data in the studies. It is known metastatic spread is strongly influenced by histology\citep{RN121} and it is conceivable that these factors are being used as surrogate markers for the histological subtype. Several studies have explored the impact of ethnicity and in recent years, mortality rates among African American have declined slower than other racial groups\citep{RN122, RN123}. The explanation is likely multifactorial with two large studies concluding that the worse survival is not due to differences in care but likely other factors such as biological differences\citep{RN125, RN126}. Therefore, while there has been no formal research linking the effectiveness of panitumumab to ethnicity, it seems plausible that such a relationship could exist but would require further research to ascertain this.

\subsection{Limitations}
Arguably, this approach has some limitations. As the methodology relies heavily on machine learning it subject to many general issues common to such techniques. Firstly, the proposed approach is computationally intensive. Secondly, the effectiveness of the approach is dependent upon sample size with large heterogenous multi-centre datasets preferred to reduce minority subgroup bias that has affected other algorithms\citep{RN96, RN95, RN56}. Thirdly, the risk of accidentally fitting confounders leading to incorrect factors being identified as important to treatment response. Undoubtedly the biggest weaknesses of the approach are common to any methods deriving subgroups post-hoc from clinical trial data. These include failure to detect the correct treatment response factors and also in identifying apparent important factors that cannot be replicated in follow-up studies\citep{RN129}. It is for this reason that results from the proposed approach should not be seen as providing definitive evidence of a subgroup but instead as suggestions for future research. A new clinical trial would be required for the findings to be truly accepted or rejected by the scientific community.

There are also a number of limitations specific to the proposed approach. In the data pre-processing stage, it is well recognised that bias could be introduced through imputing missing time series data by carrying the last value forward\citep{RN101}. More sophisticated techniques utilising functional principal component analysis could be used to address this\citep{RN102}. Further bias can result if either measurement times or censoring are not independent, as required by partly conditional methods\citep{RN103}. Notable as dependent measurement times are frequently seen in observational studies, an example being that patients tend to visit health care services more often when sick\citep{RN104}. In the model interpretability stage, the use of survLIME to characterise responders means that the proposed approach inherits its weaknesses, the most pertinent being the instability of explanations and the inability to describe non-linear relationships. The former is partially alleviated through the averaging multiple of responder explanations. Whereas the latter prevents the proposed approach from accurately describing responder subgroups defined in intervals of a certain variable. Future work could look to address this by replacing the linear surrogate Cox model used in survLIME with a non-linear survival tree. Promising work in this direction has already shown LIME to be effective when combined with a decision tree\citep{RN106}. 

\subsection{Conclusions}
In this study, a novel approach was proposed to utilise temporal information within clinical trials to better identify and characterise responders. The results suggest that this approach, specifically PCM, can improve performance compared to existing methods and yield plausible inferences when applied to clinical trial data. In the future, it is hoped that refining the approach to consider response as continuous not binary and using a tree-based surrogate model to aid interpretability may see it become an essential adjunct to drug development.

\section*{Broader Impact}
The distinctive contribution of this work is in its use of PCM to accommodate time-varying covariates thereby allowing for a dynamic response to treatment. Given the characteristics of the proposed approach, a potentially valuable application would be for use on clinical trial data during drug development. If applied throughout early and late phase clinical trials it could be used for early detection of responder subgroups which can then be tested in later confirmatory trials. Alternatively, the approach can be applied to failed late-phase clinical trials with the aim of uncovering potential subgroups of patients with positive treatment response.

\section*{Disclosure}
The authors report no conflicts of interest and declare that no funding was received for this work. Ethics approval was not required due to the exclusive use of de-identified secondary data. 

\small

\bibliographystyle{unsrtnat}
\bibliography{references}

\end{document}